\documentclass[letterpaper, 10 pt, conference]{ieeeconf}  

\IEEEoverridecommandlockouts                              

\usepackage{times}
\usepackage{epsfig}
\usepackage{graphicx}
\usepackage{amsmath}
\usepackage{amssymb}
\usepackage{subfig}
\usepackage{bbm}
\usepackage[belowskip=-15pt,aboveskip=0pt]{caption}
\usepackage{tabularx}
\usepackage{array}
\usepackage{multirow}
\usepackage{booktabs}
\usepackage{color, colortbl}
\usepackage{hyperref}

\newcommand{\fig}[1]{Fig.~\ref{#1}} 

\definecolor{Gray}{gray}{0.9}

\title{\LARGE \bf
Joint 3D Proposal Generation and Object Detection from View Aggregation
}

\author{Jason Ku, Melissa Mozifian, Jungwook Lee, Ali Harakeh, and Steven L. Waslander
\thanks{$^{1}$Jason Ku, Melissa Mozifian, Jungwook Lee, Ali Harakeh, and Steven L. Waslander are with the Department of Mechanical and Mechatronics Engineering, Faculty of Engineering, University of Waterloo, 200 University Avenue, Waterloo, ON, Canada. Email: ({\tt\small jason.ku@uwaterloo.ca, mfmozifi@uwaterloo.ca, j343lee@uwaterloo.ca, www.aharakeh.com, stevenw@uwaterloo.ca})}
}

\begin{document}
\maketitle
\thispagestyle{empty}
\pagestyle{empty}

\begin{abstract}
We present AVOD, an \textbf{A}ggregate \textbf{V}iew \textbf{O}bject \textbf{D}etection network for autonomous driving scenarios. The proposed neural network architecture uses LIDAR point clouds and RGB images to generate features that are shared by two subnetworks: a region proposal network (RPN) and a second stage detector network. The proposed RPN uses a novel architecture capable of performing multimodal feature fusion on high resolution feature maps to generate reliable 3D object proposals for multiple object classes in road scenes. Using these proposals, the second stage detection network performs accurate oriented 3D bounding box regression and category classification to predict the extents, orientation, and classification of objects in 3D space. Our proposed architecture is shown to produce state of the art results on the KITTI 3D object detection benchmark \cite{geiger2012we} while running in real time with a low memory footprint, making it a suitable candidate for deployment on autonomous vehicles. Code is at: \newline
{\url{https://github.com/kujason/avod}}

\end{abstract}

\section{Introduction}
\label{intro}
\begin{figure}[t] 
\begin{center}
\includegraphics[width=0.95\columnwidth]{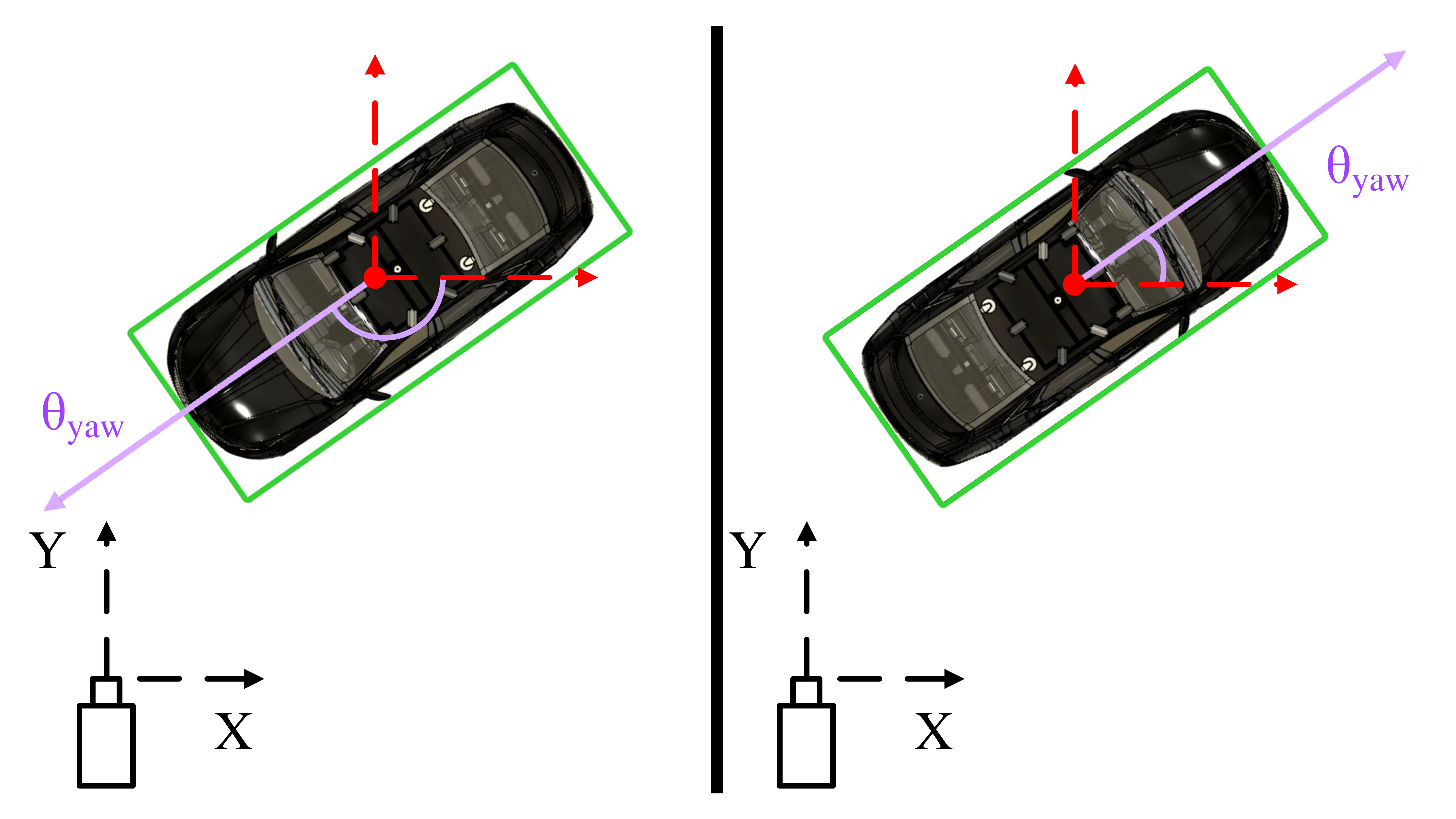}
\end{center}
\caption{A visual representation of the 3D detection problem from Bird's Eye View (BEV). The bounding box in \textbf{Green} is used to determine the IoU overlap in the computation of the \textit{average precision}. The importance of explicit orientation estimation can be seen as an object's bounding box does not change when the orientation (\textbf{purple}) is shifted by $\pm \pi$ radians.}
\label{metric}
\end{figure}

The remarkable progress made by deep neural networks on the task of 2D object detection in recent years has not transferred well to the detection of objects in 3D. The gap between the two remains large on standard benchmarks such as the KITTI Object Detection Benchmark \cite{geiger2012we} where 2D \textit{car} detectors have achieved over $90\%$ Average Precision (AP), whereas the top scoring 3D \textit{car} detector on the same scenes only achieves $70\%$ AP. The reason for such a gap stems from the difficulty induced by adding a third dimension to the estimation problem, the low resolution of 3D input data, and the deterioration of its quality as a function of distance. Furthermore, unlike 2D object detection, the 3D object detection task requires estimating \textit{oriented} bounding boxes (\fig{metric}).

Similar to 2D object detectors, most state-of-the-art deep models for 3D object detection rely on a \textit{3D region proposal generation} step for 3D search space reduction. Using region proposals allows the generation of high quality detections via more complex and computationally expensive processing at later detection stages. However, any missed instances at the proposal generation stage cannot be recovered during the following stages. Therefore, achieving a high recall during the region proposal generation stage is crucial for good performance.

Region proposal networks (RPNs) were proposed in Faster-RCNN \cite{ren2015faster}, and have become the prevailing proposal generators in 2D object detectors. RPNs can be considered a weak amodal detector, providing proposals with high recall and low precision. These deep architectures are attractive as they are able to share computationally expensive convolutional feature extractors with other detection stages. However, extending these RPNs to 3D is a non-trivial task. The Faster R-CNN RPN architecture is tailored for dense, high resolution image input, where objects usually occupy more than a couple of pixels in the feature map. When considering sparse and low resolution input such as the Front View \cite{Li-RSS-16} or Bird's Eye View (BEV) \cite{cvpr17chen} point cloud projections, this method is not guaranteed to have enough information to generate region proposals, especially for small object classes.

\begin{figure*}[t] 
\begin{center}
\includegraphics[width=0.9\textwidth]{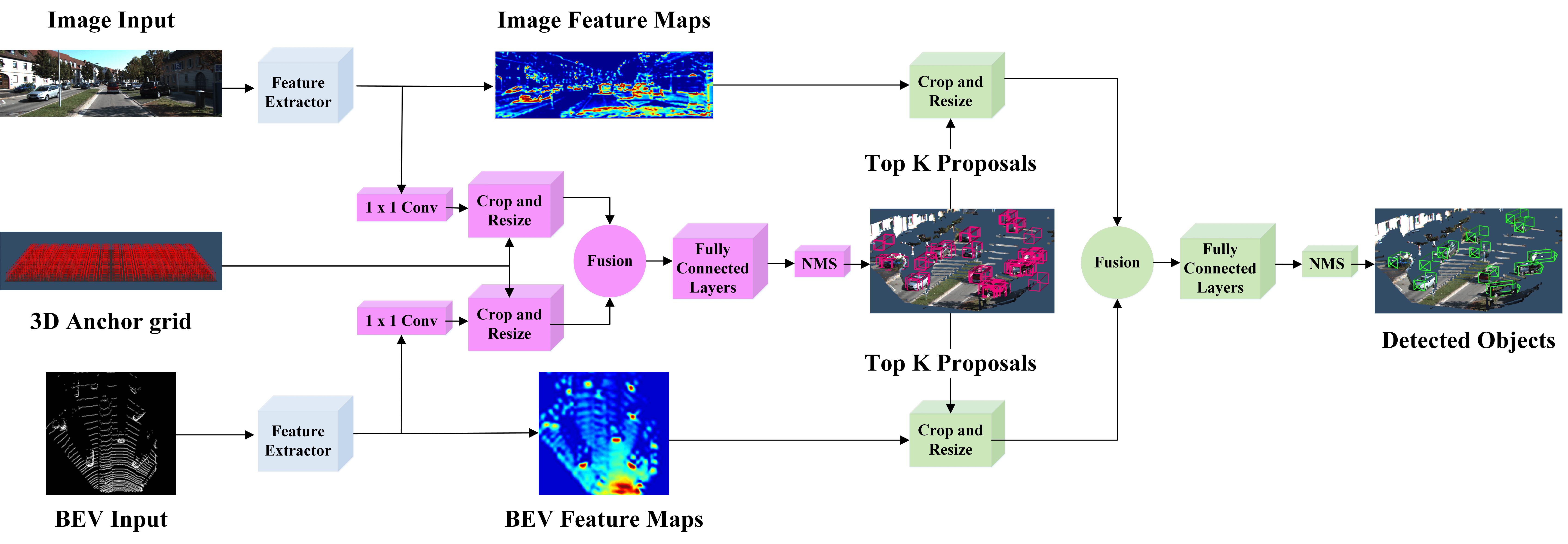}
\end{center}
\caption{The proposed method's architectural diagram. The feature extractors are shown in \textbf{blue}, the region proposal network in \textbf{pink}, and the second stage detection network in \textbf{green}.}
\label{metarch}
\end{figure*}

In this paper, we aim to resolve these difficulties by proposing AVOD, an \textbf{A}ggregate \textbf{V}iew \textbf{O}bject \textbf{D}etection architecture for autonomous driving (\fig{metarch}). The proposed architecture delivers the following contributions:
\begin{itemize}
    \item Inspired by feature pyramid networks (FPNs) \cite{lin2017feature} for 2D object detection, we propose a novel feature extractor that produces high resolution feature maps from LIDAR point clouds and RGB images, allowing for the localization of small classes in the scene.
    \item We propose a feature fusion Region Proposal Network (RPN) that utilizes multiple modalities to produce high-recall region proposals for small classes. 
    \item We propose a novel 3D bounding box encoding that conforms to box geometric constraints, allowing for higher 3D localization accuracy.
    \item The proposed neural network architecture exploits $1\times 1$ convolutions at the RPN stage, along with a fixed look-up table of 3D anchor projections, allowing high computational speed and a low memory footprint while maintaining detection performance. 
\end{itemize}

 The above contributions result in an architecture that delivers state-of-the-art detection performance at a low computational cost and memory footprint. Finally, we integrate the network into our autonomous driving stack, and show generalization to new scenes and detection under more extreme weather and lighting conditions, making it a suitable candidate for deployment on autonomous vehicles.

\section{Related Work}
\label{related}
\noindent \textbf{Hand Crafted Features For Proposal Generation:} 
Before the emergence of 3D Region Proposal Networks (RPNs) \cite{ren2015faster}, 3D proposal generation algorithms typically used hand-crafted features to generate a small set of candidate boxes that retrieve most of the objects in 3D space. 3DOP \cite{nips15chen} and Mono3D \cite{cvpr16chen} uses a variety of hand-crafted geometric features from stereo point clouds and monocular images to score 3D sliding windows in an energy minimization framework. The top $K$ scoring windows are selected as region proposals, which are then consumed by a modified Fast-RCNN \cite{girshick2015fast} to generate the final 3D detections. We use a region proposal network that learns features from both BEV and image spaces to generate higher quality proposals in an efficient manner.\\

\noindent \textbf{Proposal Free Single Shot Detectors:}
Single shot object detectors have also been proposed as RPN free architectures for the 3D object detection task. VeloFCN \cite{Li-RSS-16} projects a LIDAR point cloud to the front view, which is used as an input to a fully convolutional neural network to directly generate dense 3D bounding boxes. 3D-FCN \cite{li20163d} extends this concept by applying 3D convolutions on 3D voxel grids constructed from LIDAR point clouds to generate better 3D bounding boxes. Our two-stage architecture uses an RPN to retrieve most object instances in the road scene, providing better results when compared to both of these single shot methods. VoxelNet \cite{zhou2017voxelnet} extends 3D-FCN further by encoding voxels with point-wise features instead of occupancy values. However, even with sparse 3D convolution operations, VoxelNet's computational speed is still $3\times$ slower than our proposed architecture, which provides better results on the car and pedestrian classes.\\

\noindent \textbf{Monocular-Based Proposal Generation:}
Another direction in the state-of-the-art is using mature 2D object detectors for proposal generation in 2D, which are then extruded to 3D through amodal extent regression. This trend started with \cite{lahoud20172d} for indoor object detection, which inspired Frustum-based PointNets (F-PointNet) \cite{qi2017frustum} to use point-wise features of PointNet \cite{qi2016pointnet} instead of point histograms for extent regression. While these methods work well for indoor scenes and brightly lit outdoor scenes, they are expected to perform poorly in more extreme outdoor scenarios. Any missed 2D detections will lead to missed 3D detections and therefore, the generalization capability of such methods under such extreme conditions has yet to be demonstrated. LIDAR data is much less variable than image data and we show in Section \ref{exp} that AVOD is robust to noisy LIDAR data and lighting changes, as it was tested in snowy scenes and in low light conditions.\\

\noindent \textbf{Monocular-Based 3D Object Detectors:}
Another way to utilize mature 2D object detectors is to use prior knowledge to perform 3D object detection from monocular images only. Deep MANTA \cite{chabot2017deep} proposes a many-task vehicle analysis approach from monocular images that optimizes region proposal, detection, 2D box regression, part localization, part visibility, and 3D template prediction simultaneously. The architecture requires a database of 3D models corresponding to several types of vehicles, making the proposed approach hard to generalize to classes where such models do not exist. Deep3DBox \cite{mousavian20163d} proposes to extend 2D object detectors to 3D by exploiting the fact that the perspective projection of a 3D bounding box should fit tightly within its 2D detection window. However, in Section \ref{exp}, these methods are shown to perform poorly on the 3D detection task compared to methods that use point cloud data.\\

\noindent \textbf{3D Region Proposal Networks:}
3D RPNs have previously been proposed in \cite{song2016deep} for 3D object detection from RGBD images. However, up to our knowledge, MV3D \cite{cvpr17chen} is the only architecture that proposed a 3D RPN targeted at autonomous driving scenarios. MV3D extends the image based RPN of Faster R-CNN \cite{ren2015faster} to 3D by corresponding every pixel in the \textit{BEV} feature map to multiple prior 3D anchors. These anchors are then fed to the RPN to generate 3D proposals that are used to create view-specific feature crops from the BEV, front view of \cite{Li-RSS-16}, and image view feature maps. A \textit{deep fusion} scheme is used to combine information from these feature crops to produce the final detection output. However, this RPN architecture does not work well for small object instances in BEV. When downsampled by convolutional feature extractors, small instances will occupy a fraction of a pixel in the final feature map, resulting in insufficient data to extract informative features. Our RPN architecture aims to fuse full resolution feature crops from the image and the BEV feature maps as inputs to the RPN, allowing the generation of high recall proposals for smaller classes. Furthermore, our feature extractor provides full resolution feature maps, which are shown to greatly help in localization accuracy for small objects during the second stage of the detection framework.
\section{The AVOD Architecture}
\label{meta}
The proposed method, depicted in \fig{metarch}, uses feature extractors to generate feature maps from both the BEV map and the RGB image. Both feature maps are then used by the RPN to generate non-oriented region proposals, which are passed to the detection network for dimension refinement, orientation estimation, and category classification.

\subsection{Generating Feature Maps from Point Clouds and Images}
We follow the procedure described in \cite{cvpr17chen} to generate a six-channel BEV map from a voxel grid representation of the point cloud at $0.1$ meter resolution. The point cloud is cropped at $[-40,40]\times[0,70]$ meters to contain points within the field of view of the camera. The first $5$ channels of the BEV map are encoded with the maximum height of points in each grid cell, generated from $5$ equal slices between $[0,2.5]$ meters along the Z axis. The sixth BEV channel contains point density information computed per cell as $\min(1.0, \frac{\log(N+1)}{\log 16})$, where $N$ is the number of points in the cell.

\subsection{The Feature Extractor}
\label{extract}
\begin{figure}[t] 
\centering
\includegraphics[width= 0.8\columnwidth]{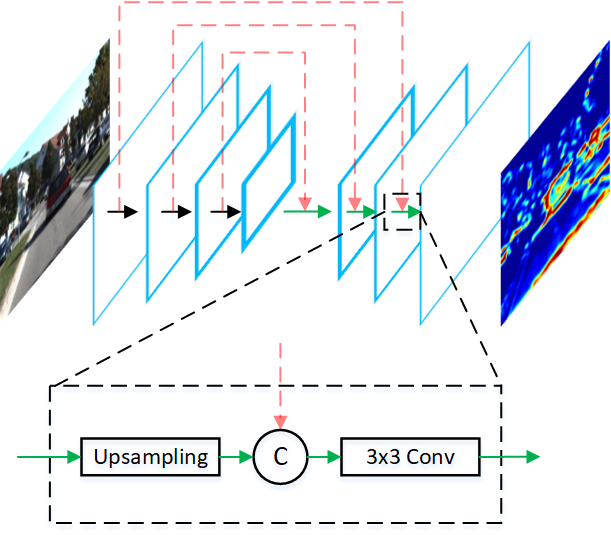}
\caption{The architecture of our proposed high resolution feature extractor shown here for the image branch. Feature maps are propagated from the encoder to the decoder section via red arrows. Fusion is then performed at every stage of the decoder by a learned upsampling layer, followed by concatenation, and then mixing via a convolutional layer, resulting in a full resolution feature map at the last layer of the decoder.}
\label{pyr}
\end{figure}

The proposed architecture uses two identical feature extractor architectures, one for each input view. The full-resolution feature extractor is shown in \fig{pyr} and is comprised of two segments: an encoder and a decoder. The encoder is modeled after VGG-16 \cite{simonyan2014very} with some modifications, mainly a reduction of the number of channels by half, and cutting the network at the conv-4 layer. The encoder therefore takes as an input an $M \times N \times D$ image or BEV map, and produces an $\frac{M}{8}\times \frac{N}{8} \times D^*$ feature map $F$. $F$ has high representational power, but is $8\times$ lower in resolution compared to the input. An average pedestrian in the KITTI dataset occupies $0.8\times0.6$ meters in the BEV. This translates to an $8\times6$ pixel area in a BEV map with $0.1$ meter resolution. Downsampling by $8\times$ results in these small classes to occupy \textit{less than one} pixel in the output feature map, that is without taking into account the increase in receptive field caused by convolutions. Inspired by the Feature Pyramid Network (FPN) \cite{lin2017feature}, we create a bottom-up decoder that learns to upsample the feature map back to the original input size, while maintaining run time speed. The decoder takes as an input the output of the encoder, $F$, and produces a new $M \times N \times \tilde{D}$ feature map. \fig{pyr} shows the operations performed by the decoder, which include upsampling of the input via a \textit{conv-transpose} operation, concatenation of a corresponding feature map from the encoder, and finally fusing the two via a $3\times3$ convolution operation. The final feature map is of high resolution \textit{and} representational power, and is shared by both the RPN and the second stage detection network. 

\subsection{Multimodal Fusion Region Proposal Network}
Similar to 2D two-stage detectors, the proposed RPN regresses the difference between a set of prior 3D boxes and the ground truth. These prior boxes are referred to as anchors, and are encoded using the \textit{axis aligned} bounding box encoding shown in \fig{encoding}. Anchor boxes are parameterized by the centroid $(t_x, t_y, t_z)$ and axis aligned dimensions $(d_x, d_y, d_z)$. To generate the 3D anchor grid, $(t_x, t_y)$ pairs are sampled at an interval of $0.5$ meters in BEV, while $t_z$ is determined based on the sensor's height above the ground plane. The dimensions of the anchors are determined by clustering the training samples for each class. Anchors without 3D points in BEV are removed efficiently via integral images resulting in $80-100$K non-empty anchors per frame.
\\
\\
\textbf{Extracting Feature Crops Via Multiview Crop And Resize Operations:}
To extract feature crops for every anchor from the view specific feature maps, we use the crop and resize operation \cite{Huang_2017_CVPR}. Given an anchor in 3D, two regions of interest are obtained by projecting the anchor onto the BEV and image feature maps. The corresponding regions are then used to extract feature map crops from each view, which are then bilinearly resized to $3\times3$ to obtain equal-length feature vectors. This extraction method results in feature crops that abide by the aspect ratio of the projected anchor in both views, providing a more reliable feature crop than the $3\times3$ convolution used originally by Faster-RCNN.
\\
\\ 
\textbf{Dimensionality Reduction Via $1\times1$ Convolutional Layers:} In some scenarios, the region proposal network is required to save feature crops for $100$K anchors in GPU memory. Attempting to extract feature crops directly from high dimensional feature maps imposes a large memory overhead per input view. As an example, extracting $7\times7$ feature crops for $100$K anchors from a $256$-dimensional feature map requires around $5$ gigabytes\footnote{$100,000\times 7\times 7\times 256\times 4$ bytes.} of memory assuming $32$-bit floating point representation. Furthermore, processing such high-dimensional feature crops with the RPN greatly increases its computational requirements.

Inspired by their use in \cite{iandola2016squeezenet}, we propose to apply a $1\times 1$ convolutional kernel on the output feature maps from each view, as an efficient dimensionality reduction mechanism that learns to select features that contribute greatly to the performance of the region proposal generation. This reduces the memory overhead for computing anchor specific feature crops by $\tilde{D} \times$, allowing the RPN to process fused features of tens of thousands of anchors using only a few megabytes of additional memory.
\\
\\
\textbf{3D Proposal Generation:} The outputs of the crop and resize operation are equal-sized feature crops from both views, which are fused via an element-wise mean operation. Two task specific branches \cite{ren2015faster} of fully connected layers of size $256$ use the fused feature crops to regress axis aligned object proposal boxes and output an object/background ``objectness'' score. 3D box regression is performed by computing $(\Delta t_x, \Delta t_y, \Delta t_z, \Delta d_x, \Delta d_y, \Delta d_z)$, the difference in centroid and dimensions between anchors and ground truth bounding boxes. Smooth L1 loss is used for 3D box regression, and cross-entropy loss for ``objectness''. Similar to \cite{ren2015faster}, background anchors are ignored when computing the regression loss. Background anchors are determined by calculating the 2D IoU in BEV between the anchors and the ground truth bounding boxes. For the \textit{car} class, anchors with IoU less than $0.3$ are considered background anchors, while ones with IoU greater than $0.5$ are considered object anchors. For the \textit{pedestrian} and \textit{cyclist} classes, the object anchor IoU threshold is reduced to $0.45$. To remove redundant proposals, 2D non-maximum suppression (NMS) at an IoU threshold of $0.8$ in BEV is used to keep the top $1024$ proposals during training. At inference time, $300$ proposals are used for the car class, whereas $1024$ proposals are kept for pedestrians and cyclists.

\subsection{Second Stage Detection Network}
\label{seco}
\noindent \textbf{3D Bounding Box Encoding:} In \cite{cvpr17chen}, Chen et al. claim that 8 corner box encoding  provides better results than the traditional axis aligned encoding previously proposed in \cite{song2016deep}. However, an 8 corner encoding does not take into account the physical constraints of a 3D bounding box, as the top corners of the bounding box are forced to align with those at the bottom. To reduce redundancy and keep these physical constraints, we propose to encode the bounding box with four corners and two height values representing the top and bottom corner offsets from the ground plane, determined from the sensor height. Our regression targets are therefore $(\Delta x_1 ... \Delta x_4, \Delta y_1 ... \Delta y_4, \Delta h_1 , \Delta h_2$), the corner and height offsets from the ground plane between the proposals and the ground truth boxes. To determine corner offsets, we correspond the closest corner of the proposals to the closest corner of the ground truth box in BEV. The proposed encoding reduces the box representation from an overparameterized $24$ dimensional vector to a $10$ dimensional one. \\

\noindent\textbf{Explicit Orientation Vector Regression:} To determine orientation from a 3D bounding box, MV3D \cite{cvpr17chen} relies on the extents of the estimated bounding box where the orientation vector is assumed to be in the direction of the longer side of the box. This approach suffers from two problems. First, this method fails for detected objects that do not always obey the rule proposed above, such as pedestrians. Secondly, the resulting orientation is only known up to an additive constant of $\pm \pi$ radians. Orientation information is lost as the corner order is not preserved in closest corner to corner matching. \fig{metric} presents an example of how the same rectangular bounding box can contain two instances of an object with opposite orientation vectors. Our architecture remedies this problem by computing $(x_{\theta}, y_{\theta})= (\cos(\theta),\sin(\theta))$. This orientation vector representation implicitly handles angle wrapping as every $\theta \in [-\pi,\pi]$ can be represented by a unique unit vector in the BEV space. We use the regressed orientation vector to \textit{resolve} the ambiguity in the bounding box orientation estimate from the adopted four corner representation, as this experimentally found to be more accurate than using the regressed orientation directly. Specifically, we extract the four possible orientations of the bounding box, and then choose the one closest to the explicitly regressed orientation vector.
\\
\\
\textbf{Generating Final Detections:}
Similar to the RPN, the inputs to the multiview detection network are feature crops generated from projecting the proposals into the two input views. As the number of proposals is an order of magnitude lower than the number of anchors, the original feature map with a depth of $\tilde{D}=32$ is used for generating these feature crops. Crops from both input views are resized to $7\times7$ and then fused with an element-wise mean operation. A \textit{single} set of three fully connected layers of size $2048$ process the fused feature crops to output box regression, orientation estimation, and category classification for each proposal. Similar to the RPN, we employ a multi-task loss combining two Smooth L1 losses for the bounding box and orientation vector regression tasks, and a cross-entropy loss for the classification task. Proposals are only considered in the evaluation of the regression loss if they have at least a $0.65$ or $0.55$ 2D IoU in BEV with the ground truth boxes for the \textit{car} and \textit{pedestrian/cyclist} classes, respectively. To remove overlapping detections, NMS is used at a threshold of $0.01$.

\subsection{Training}
We train two networks, one for the \textit{car} class and one for both the \textit{pedestrian} and \textit{cyclist} classes. The RPN and the detection networks are trained jointly in an end-to-end fashion using mini-batches containing one image with $512$ and $1024$ ROIs, respectively. The network is trained for $120$K iterations using an ADAM optimizer with an initial learning rate of $0.0001$ that is decayed exponentially every $30$K iterations with a decay factor of $0.8$.  

\begin{figure}[t] 
\begin{center}
\includegraphics[width=0.45\textwidth]{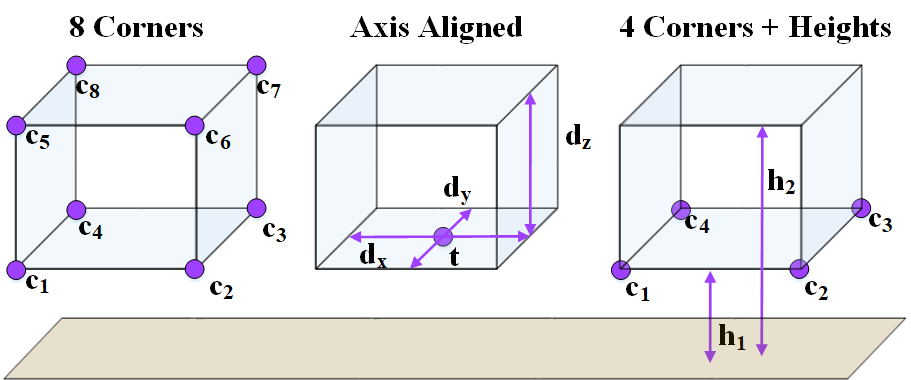}
\end{center}
\caption{A visual comparison between the $8$ corner box encoding proposed in \cite{cvpr17chen}, the axis aligned box encoding proposed in \cite{song2016deep}, and our $4$ corner encoding.}
\label{encoding}
\end{figure}

\begin{figure*}[t] 
\begin{center}
\includegraphics[width=0.99\textwidth]{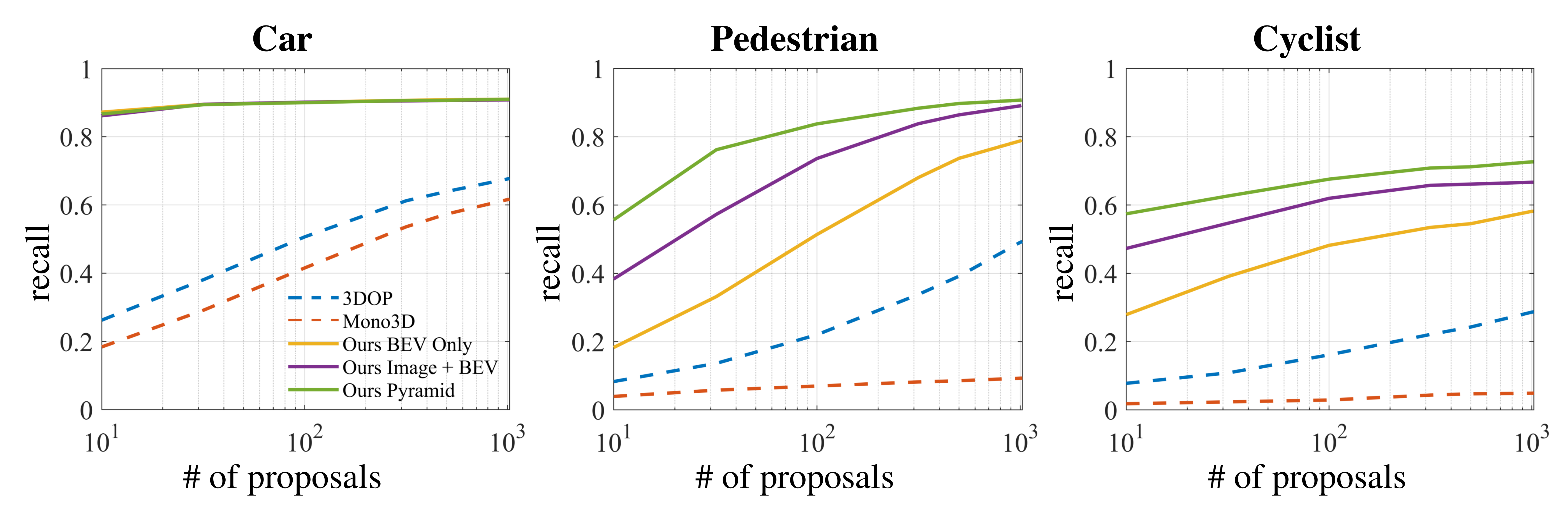}
\end{center}
\caption{Recall vs. number of proposals at a 3D IoU threshold of 0.5 for the three classes evaluated on the \textit{validation} set at \textit{moderate} difficulty.}
\label{rpn_res}
\end{figure*}

\section{Experiments and Results}
\label{exp}
We test AVOD's performance on the proposal generation and object detection tasks on the three classes of the KITTI Object Detection Benchmark \cite{geiger2012we}. We follow \cite{cvpr17chen} to split the provided $7481$ training frames into a \textit{training} and a \textit{validation} set at approximately a $1:1$ ratio. For evaluation, we follow the \textit{easy, medium, hard} difficulty classification proposed by KITTI. We evaluate and compare two versions of our implementation, \textit{Ours} with a VGG-like feature extractor similar to \cite{cvpr17chen}, and \textit{Ours (Feature Pyramid)} with the proposed  high resolution feature extractor described in Section \ref{extract}.
\\
\\
\textbf{3D Proposal Recall:}
3D proposal generation is evaluated using \textit{3D bounding box recall} at a $0.5$ 3D IoU threshold. We compare three variants of our RPN against the proposal generation algorithms 3DOP \cite{nips15chen} and Mono3D \cite{cvpr16chen}. \fig{rpn_res} shows the recall vs number of proposals curves for our RPN variants, 3DOP and Mono3D. It can be seen that our RPN variants outperform both 3DOP and Mono3D by a wide margin on all three classes. As an example, our Feature Pyramid based fusion RPN achieves an \textbf{$86\%$} 3D recall on the \textit{car} class with just 10 proposals per frame. The maximum recall achieved by 3DOP and Mono3D on the \textit{car} class is $73.87\%$ and $65.74\%$ respectively. This gap is also present for the pedestrian and cyclist classes, where our RPN achieves more than $20\%$ increase in recall at $1024$ proposals. This large gap in performance suggests the superiority of learning based approaches over methods based on hand crafted features. For the car class, our RPN variants achieve a $91\%$ recall at just $50$ proposals, whereas MV3D \cite{cvpr17chen} reported requiring $300$ proposals to achieve the same recall. It should be noted that MV3D does not publicly provide proposal results for cars, and was not tested on pedestrians or cyclists.
\\
\begin{table}[b]
\centering
\resizebox{0.99\columnwidth}{!}{
\begin{tabular}{c c c c c c c}
&\multicolumn{2}{c}{Easy} &\multicolumn{2}{c}{Moderate} & \multicolumn{2}{c}{Hard}\\
\cmidrule{2-7}
\cmidrule{2-7}
& AP & AHS & AP & AHS & AP & AHS  \\
\midrule
Deep3DBox & 5.84 & 5.84 & 4.09 & 4.09 & 3.83 & 3.83  \\
MV3D & 83.87 & 52.74 & 72.35 & 43.75 & 64.56 & 39.86 \\
Ours (Feature Pyramid) & \textbf{84.41} & \textbf{84.19} & \textbf{74.44} & \textbf{74.11}& \textbf{68.65} & \textbf{68.28} \\ 
\bottomrule   
\end{tabular}
}
\caption{A comparison of the performance of Deep3DBox \cite{mousavian20163d}, MV3D \cite{cvpr17chen}, and our method evaluated on the \textit{car} class in the \textit{validation} set. For evaluation, we show the AP and AHS (in $\%$) at $0.7$ 3D IoU.}
\label{comparison_val}
\end{table}

\begin{table*}[t]
\centering
\resizebox{0.9\textwidth}{!}{
\begin{tabular}{ccc|ccc|ccc}
\multicolumn{3}{c}{} & \multicolumn{3}{c}{$AP_{3D} \ (\%)$} & \multicolumn{3}{c}{$AP_{BEV} (\%)$} \\
\midrule
Method & Runtime (s) & Class & Easy & Moderate & Hard & Easy & Moderate & Hard \\
\midrule
MV3D \cite{cvpr17chen}& $0.36$ &\multirow{5}{*}{\textbf{Car}} & 71.09 & 62.35 & 55.12 & 86.02 & 76.90 & 68.49 \\
VoxelNet \cite{zhou2017voxelnet}& $0.23$ & &77.47 & 65.11 & 57.73 & \textbf{89.35} & 79.26 & 77.39 \\
F-PointNet \cite{qi2017frustum}& $0.17$ & & 81.20 & 70.39 & 62.19 & 88.70 & 84.00 & 75.33 \\
Ours & \textbf{0.08} & & 73.59 & 65.78 & 58.38 & 86.80 & \textbf{85.44} & 77.73\\
Ours (Feature Pyramid)& 0.1 & & \textbf{81.94} & \textbf{71.88} & \textbf{66.38} & 88.53 & 83.79 & \textbf{77.90} \\
\midrule
VoxelNet \cite{zhou2017voxelnet} & $0.23$ & \multirow{4}{*}{\textbf{Ped.}}& 39.48 & 33.69 & 31.51 & 46.13 & 40.74 & 38.11 \\
F-PointNet  \cite{qi2017frustum} & $0.17$ & & \textbf{51.21} & \textbf{44.89} & 40.23 & 58.09 & 50.22 & 47.20 \\
Ours & \textbf{0.08} & & 38.28 & 31.51 & 26.98 & 42.51 & 35.24 & 33.97 \\
Ours (Feature Pyramid)& $0.1$ & & 50.80 & 42.81 & \textbf{40.88} & \textbf{58.75} & \textbf{51.05} & \textbf{47.54} \\

\midrule
VoxelNet \cite{zhou2017voxelnet} & $0.23$ & \multirow{4}{*}{\textbf{Cyc.}} & 61.22 & 48.36  & 44.37 & 66.70 & 54.76 & 50.55 \\
F-PointNet  \cite{qi2017frustum} & $0.17$ & & \textbf{71.96} & \textbf{56.77}& \textbf{50.39} & \textbf{75.38} & \textbf{61.96} & \textbf{54.68} \\
Ours & \textbf{0.08} & & 60.11 & 44.90& 38.80 & 63.66  & 47.74  & 46.55 \\
Ours (Feature Pyramid)& $0.1$ & & 64.00  & 52.18 & 46.61 &68.06 & 57.48  & 50.77 \\
\bottomrule   
\end{tabular}
}
\caption{A comparison of the performance of AVOD with the state of the art 3D object detectors evaluated on KITTI's \textit{test} set. Results are generated by KITTI's evaluation server \cite{kitti}.}
\label{comparison_test}
\end{table*}

\begin{figure}[t] 
\begin{center}
\includegraphics[width=0.9\columnwidth]{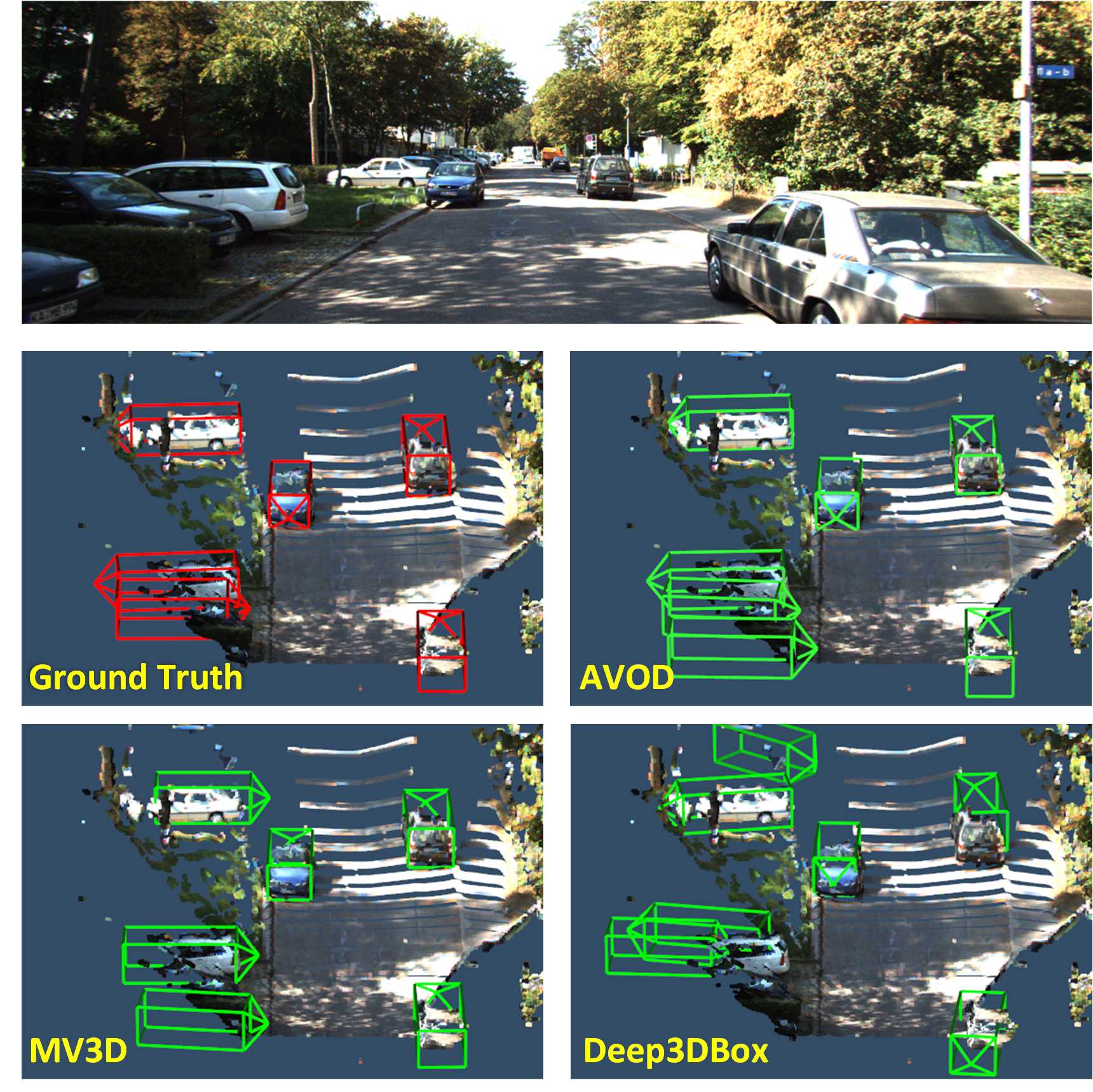}
\end{center}
\caption{A qualitative comparison between MV3D \cite{cvpr17chen}, Deep3DBox \cite{mousavian20163d}, and our architecture relative to KITTI's ground truth on a sample in the \textit{validation} set.}
\label{res_comp}
\end{figure}

\noindent\textbf{3D Object Detection}
3D detection results are evaluated using the 3D and BEV AP and Average Heading Similarity (AHS) at $0.7$ IoU threshold for the car class, and $0.5$ IoU threshold for the pedestrian and cyclist classes. The AHS is the Average Orientation Similarity (AOS) \cite{geiger2012we}, but evaluated using 3D IOU and global orientation angle instead of $2D$ IOU and observation angle, removing the metric's dependence on localization accuracy. We compare against publicly provided detections from MV3D \cite{cvpr17chen} and Deep3DBox \cite{mousavian20163d} on the \textit{validation} set. It has to be noted that no currently published method publicly provides results on the pedestrian and cyclist classes for the 3D object detection task, and hence comparison is done for the \textit{car} class only. On the \textit{validation} set (Table \ref{comparison_val}), our architecture is shown to outperform MV3D by $2.09\%$ AP on the \textit{moderate} setting and $4.09\%$ on the \textit{hard} setting. However, AVOD achieves a $30.36\%$ and $28.42\%$ increase in AHS over MV3D at the \textit{moderate} and \textit{hard} setting respectively. This can be attributed to the loss of orientation vector direction discussed in Section \ref{seco} resulting in orientation estimation up to an additive error of $\pm \pi$ radians. To verify this assertion, \fig{res_comp} shows a visualization of the results of AVOD and MV3D in comparison to KITTI's ground truth. 
It can be seen that MV3D assigns erroneous orientations for almost half of the cars shown. On the other hand, our proposed architecture assigns the correct orientation for all cars in the scene. As expected, the gap in 3D localization performance between Deep3DBox and our proposed architecture is very large. It can be seen in \fig{res_comp} that Deep3DBox fails at accurately localizing most of the vehicles in 3D. This further enforces the superiority of fusion based methods over monocular based ones. We also compare the performance of our architecture on the KITTI \textit{test} set with MV3D, VoxelNet\cite{zhou2017voxelnet}, and F-PointNet\cite{qi2017frustum}. \textit{Test} set results are provided directly by the evaluation server, which does not compute the AHS metric. Table \ref{comparison_test} shows the results of AVOD on KITTI's \textit{test} set. It can be seen that even with only the encoder for feature extraction, our architecture performs quite well on all three classes, while being \textit{twice} as fast as the next fastest method, F-PointNet. However, once we add our high-resolution feature extractor (Feature Pyramid), our architecture outperforms all other methods on the car class in 3D object detection, with a noticeable margin of $4.19\%$ on hard (highly occluded or far) instances in comparison to the second best performing method, F-PointNet. On the pedestrian class, our Feature Pyramid architecture ranks \textbf{first} in BEV AP, while scoring slightly above F-PointNet on hard instances using 3D AP. On the cyclist class, our method falls short to F-PointNet. We believe that this is due to the low number of cyclist instances in the KITTI dataset, which induces a bias towards pedestrian detection in our joint pedestrian/cyclist network. \\

\begin{figure*}[t] 
\begin{center}
\includegraphics[width=0.99\textwidth]{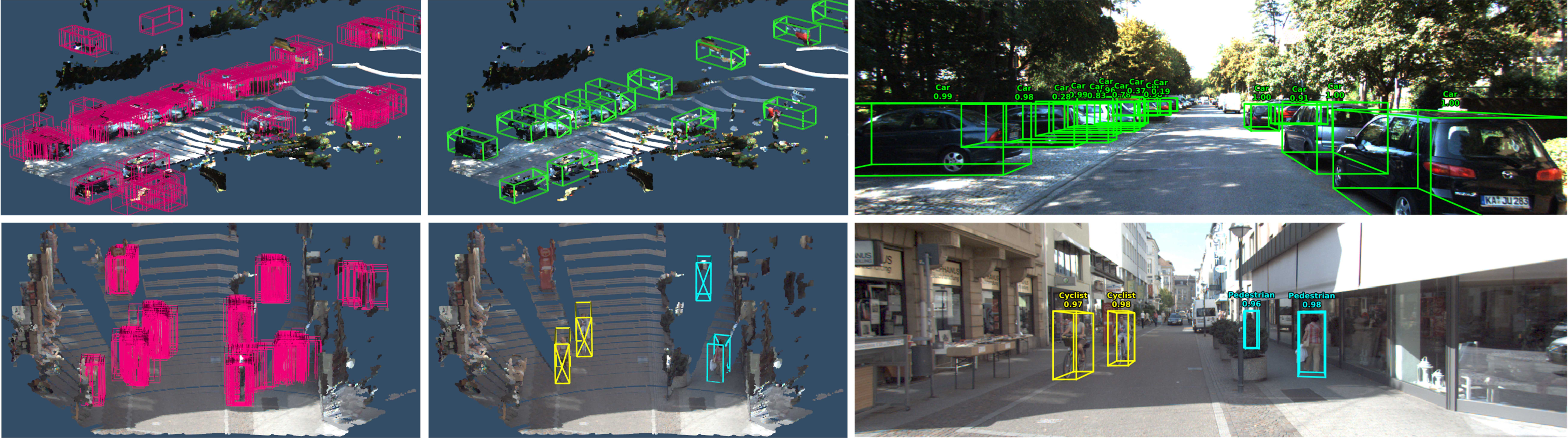}
\end{center}
\caption{Qualitative results of AVOD for cars (top) and pedestrians/cyclists (bottom). \textbf{Left:} 3D region proposal network output, \textbf{Middle:} 3D detection output, and \textbf{Right:} the projection of the detection output onto image space for all three classes. The 3D LIDAR point cloud has been colorized and interpolated for better visualization. \\}
\label{res_qual}
\end{figure*}

\noindent\textbf{Runtime and Memory Requirements:} We use FLOP count and number of parameters to assess the computational efficiency and the memory requirements of the proposed network. Our final Feature Pyramid fusion architecture employs roughly $38.073$ million parameters, approximately $16\%$ that of MV3D. The \textit{deep fusion} scheme employed by MV3D triples the number of fully connected layers required for the second stage detection network, which explains the significant reduction in the number of parameters by our proposed architecture. Furthermore, our Feature Pyramid fusion architecture requires $231.263$ billion FLOPs per frame allowing it to process frames in $0.1$ seconds on a TITAN Xp GPU, taking 20ms for pre-processing and 80ms for inference. This makes it $1.7\times$ faster than F-PointNet, while maintaining state-of-the-art results. Finally, our proposed architecture requires only $2$ gigabytes of GPU memory at inference time, making it suitable to be used for deployment on autonomous vehicles.

\subsection{Ablation Studies:}
Table \ref{hyperparam} shows the effect of varying different hyperparameters on the performance measured by the AP and AHS, number of model parameters, and FLOP count of the proposed architecture. The base network uses hyperparameter values described throughout the paper up to this point, along with the feature extractor of MV3D. We study the effect of the RPN's input feature vector origin and size on both the proposal recall and final detection AP by training two networks, one using \textit{BEV only} features and the other using feature crops of size $1\times1$ as input to the RPN stage. We also study the effect of different bounding box encoding schemes shown in \fig{encoding}, and the effects of adding an orientation regression output layer on the final detection performance in terms of AP and AHS. Finally, we study the effect of our high-resolution feature extractor, compared to the original one proposed by MV3D.\\

\noindent \textbf{RPN Input Variations:} \fig{rpn_res} shows the recall vs number of proposals curves for both the original RPN and BEV only RPN \textit{without the feature pyramid extractor} on the three classes on the \textit{validation} set. For the pedestrian and cyclist classes, fusing features from both views at the RPN stage is shown to provide a \textbf{$10.1\%$} and\textbf{ $8.6\%$} increase in recall over the BEV only version at $1024$ proposals.  Adding our high-resolution feature extractor increases this difference to around \textbf{$10.5 \%$} and \textbf{$10.8 \%$} for the respective classes. For the car class, adding image features as an input to the RPN, or using the high resolution feature extractor does not seem to provide a higher recall value over the BEV only version. We attribute this to the fact that instances from the car class usually occupy a large space in the input BEV map, providing sufficient features in the corresponding output low resolution feature map to reliably generate object proposals. The effect of the increase in proposal recall on the final detection performance can be observed in Table \ref{hyperparam}. Using both image and BEV features at the RPN stage results in a $6.9\%$ and $9.4\%$ increase in AP over the BEV only version for the pedestrian and cyclist classes respectively.\\ 

\begin{table*}[t]
\centering
\resizebox{0.95\textwidth}{!}{
\begin{tabular}{c c c c c c c c c}
\toprule
&\multicolumn{2}{c}{Car}&\multicolumn{2}{c}{Pedestrian}&\multicolumn{2}{c}{Cyclist}& &\\ \cmidrule{2-3} \cmidrule{4-5} \cmidrule{6-7}
Architecture & AP & AHS & AP & AHS & AP & AHS & Number Of Parameters & FLOPs\\
\midrule
\midrule
\rowcolor{Gray}
Base Network &  74.1 & 73.9 & 39.5 & 29.8 & 41.6 & 33.2 & 38,073,528 & 186,284,945,569 \\

RPN BEV Only & 74.1 & 73.9 & 32.6 & 26.7 & 32.2 & 30.2 & 0 & \textbf{-266,641,569} \\

Axis-Aligned Boxes & 67.6 & 67.5 & 36.2 & 27.8 & 36.6 & 35.7& -8196 & 0 \\

Non-Ordered 4 Corners, No Orientation & 67.8 & 43.1 & 37.8 & 17.9 & 41.1 & 18.6 & -4,098 & -20,418 \\ 

Non-Ordered 8 Corners, No Orientation & 66.9 & 34.1 & 37.8 & 18.1 & 40.4 & 21.0 & +24,638 & +122,879 \\ 

Feature Pyramid Extractor & \textbf{74.4} & \textbf{74.1}& \textbf{58.8}& \textbf{43.3}& \textbf{49.7}& \textbf{48.7}& \textbf{-21,391,104} & +44,978,386,776 \\

\bottomrule   
\end{tabular}
}
\caption{A comparison of the performance of different variations of hyperparameters, evaluated on the \textit{validation} set at \textit{moderate} difficulty. We use a 3D IoU threshold of $0.7$ for the Car class, and $0.5$ for the pedestrian and cyclist classes. The effect of variation of hyperparameters on the FLOPs and number of parameters are measured relative to the base network.}
\label{hyperparam}
\end{table*}

\noindent \textbf{Bounding Box Encoding:} 
We study the effect of different bounding box encodings shown in \fig{encoding} by training two additional networks. The first network estimates axis aligned bounding boxes, using the regressed orientation vector as the final box orientation. The second and the third networks use our $4$ corner and MV3D's $8$ corner encodings without additional orientation estimation as described in Section \ref{seco}. As expected, without orientation regression to provide orientation angle correction, the two networks employing the $4$ corner and the $8$ corner encodings provide a much lower AHS than the base network for all three classes. This phenomenon can be attributed to the loss of orientation information as described in Section \ref{seco}.\\

\noindent \textbf{Feature Extractor:}
We compare the detection results of our feature extractor to that of the base VGG-based feature extractor proposed by MV3D. For the \textit{car} class, our pyramid feature extractor only achieves a gain of $0.3\%$ in AP and AHS. However, the performance gains on smaller classes is much more substantial. Specifically, we achieve a gain of $19.3\%$ and $8.1\%$ AP on the \textit{pedestrian} and \textit{cyclist} classes respectively. This shows that our high-resolution feature extractor is essential to achieve state-of-the-art results on these two classes with a minor increase in computational requirements.\\ 

\noindent\textbf{Qualitative Results:} \fig{res_qual} shows the output of the RPN and the final detections in both 3D and image space. More qualitative results including those of AVOD running in \textbf{snow} and \textbf{night} scenes are provided at \href{https://youtu.be/mDaqKICiHyA}{https://youtu.be/mDaqKICiHyA}.

\section{Conclusion}
In this work we proposed AVOD, a 3D object detector for autonomous driving scenarios. The proposed architecture is differentiated from the state of the art by using a high resolution feature extractor coupled with a multimodal fusion RPN architecture, and is therefore able to produce accurate region proposals for small classes in road scenes. Furthermore, the proposed architecture employs explicit orientation vector regression to resolve the ambiguous orientation estimate inferred from a bounding box. Experiments on the KITTI dataset show the superiority of our proposed architecture over the state of the art on the 3D localization, orientation estimation, and category classification tasks. Finally, the proposed architecture is shown to run in real time and with a low memory overhead.

\bibliography{root}
\end{document}